\documentclass[11pt]{article}
\usepackage[utf8]{inputenc}
\usepackage[T1]{fontenc}
\usepackage{multirow} 
\usepackage{longtable} 
\usepackage{array}
\usepackage{makecell}
\usepackage{hyperref}
\usepackage{amsmath} 
\usepackage{graphicx} 
\DeclareUnicodeCharacter{2082}{\textsubscript{2}}

\title{Brick Kiln Dataset for Pakistan's IGP Region Using AI}

\title{Brick Kiln Dataset for Pakistan's IGP Region Using AI}

\author{Muhammad Suleman Ali Hamdani$^1$, Khizer Zakir$^2$, Neetu Kushwaha$^3$, \\ Syeda Eman Fatima$^1$,  Hassan Aftab Sheikh$^3$}

\date{}

\begin{document}

\maketitle

\begin{center}
    \small
    $^1$School of Electrical Engineering and Computer Science, National University of Sciences and Technology, Islamabad, 46000, Pakistan \\
    $^2$Department of Geoinformatics – Z\_GIS, University of Salzburg, 5020, Austria \\
    $^3$Smith School of Enterprise and Environment, University of Oxford, South Parks Rd, Oxford OX1 3QY \\
\end{center}

\begin{abstract}
\noindent Brick kilns are a major source of air pollution in Pakistan, with many operating without regulation. A key challenge in Pakistan and across the Indo-Gangetic Plain is the limited air quality monitoring and lack of transparent data on pollution sources. To address this, we present a two-fold AI approach that combines low-resolution Sentinel-2 and high-resolution imagery to map brick kiln locations. Our process begins with a low-resolution analysis, followed by a post-processing step to reduce false positives, minimizing the need for extensive high-resolution imagery. This analysis initially identified 20,000 potential brick kilns, with high-resolution validation confirming around 11,000 kilns. The dataset also distinguishes between Fixed Chimney and Zigzag kilns, enabling more accurate pollution estimates for each type. Our approach demonstrates how combining satellite imagery with AI can effectively detect specific polluting sources. This dataset provides regulators with insights into brick kiln pollution, supporting interventions for unregistered kilns and actions during high pollution episodes.\\
\end{abstract}

\flushbottom
\maketitle

\thispagestyle{empty}

\section*{Background \& Summary}

Air pollution in South Asia is responsible for over 2 million premature deaths annually with pollution levels  exceeding World Health Organization (WHO) air quality standards by up to ten times for particulate matter (PM$_{2.5}$) \cite{worldbank2023, iqair2023mostpolluted}. The Indo-Gangetic Plain (IGP), a region (Figure \ref{fig:igp}a) encompassing 13.5 million hectares of land across India, Pakistan, and Bangladesh is one of the most polluted regions globally \cite{shi2020urbanization,teri2019scoping}. The region's meteorological and topographical region restricts the dispersion of pollutants, making it a global air pollution hotspot {\cite{khan_critical_2024}. This is also highlighted from satellite data observations that have shown a temporal increase from 2011 to 2021 in pollutants such as sulphur dioxide (SO$_2$), nitrogen oxides (NO$_x$), and other aerosols \cite{rahman2023evaluation}. \\

\noindent In developing countries, air pollution from informal sectors is often unregulated and can go unabated due to limited or no regulations. This is partly due to little or no data available \cite{mazhar2013}. In the IGP region, the sources of air pollution include industrial activities \cite{sumaira2022industrialization}, vehicular emissions \cite{jabbar2022air}, burning of agricultural crop residue \cite{ravindra2021appraisal}, and transfer of trans-boundary pollutants. In particular, trans-boundary sources of air pollutants can impact air quality across countries as has been observed in the IGP with India being the major source of pollution affecting both Pakistan and Bangladesh. The latter is particularly vulnerable, where most air masses from the Indian IGP region transports pollutants across borders \cite{rana2016transboundary} contributing up to 40\% of pollution in Bangladesh-India bordering cities \cite{dibya2023}. \\

\noindent In South Asia, the brick kiln industry contributes not just to greenhouse gas emissions (GHG) but also to non-GHG emissions. Asian continent contributes to 87\% of the total 1.5 trillion clay bricks manufactured annually with South-Asian countries (Pakistan, India, Nepal, and Bangladesh) producing 20\% of those second only to China \cite{seay2021impact}. The kilns in South Asia are fueled by low- to medium-grade coal. Non-GHG emissions associated with producing bricks include  particulate matter (PM), sulphur dioxide (SO$_2$), nitrogen oxides (NO$_x$), and carbon monoxide (CO) emissions pose a health hazard. A World Bank report claims that the brick kiln industry can be responsible for up to 91\% of air pollution in some cities of the three South Asian countries mentioned\cite{eil2020dirty}. \\

\noindent The short- and long-term exposure to the listed pollutants poses a health risk. Fine particulate matter such as PM$_{2.5}$ or smaller can penetrate deep into the lungs, circular system, and other organs \cite{pryor2022physiological}. Exposure to PM$_{2.5}$ can cause cardiovascular, respiratory, and pulmonary diseases with the potential of being fatal \cite{guo2023longterm}. Research has also shown evidence of air pollution leading to epigenetic modifications which can be inherited across generations \cite{shukla2019air}. This highlights the potential impacts of air pollution on the neurodevelopmental growth of future generations. Studies have shown a link between PM, CO, NO$_{2}$, and SO$_{2}$ exposure during pregnancy and an increased probability of premature births and low birth weights in infants \cite{bazyar2019ambient, who2023typesofpollutants} . This is a critical issue, given Pakistan has one of the highest fertility rates in South Asia \cite{worldbank2022fertilityrate}.
Similarly, prolonged exposure to air pollution significantly reduces the life expectancy, such as 4.3 years in Pakistan\cite{greenstone2023pakistan}, 5.3 years in India by \cite{greenstone2023india}, and a staggering 6.8 years in Bangladesh\cite{greenstone2023bangladesh} on average.  \\

\noindent Despite knowing the health hazards of air pollution, a major challenge remains: accurately monitoring the impacts of various pollutants, identifying and understanding their sources. \\

\noindent The lack of a publicly available database for brick kilns in Pakistan also correlates with the informal and often illegal operation of the facility. The kilns in South Asia are outdated but research suggests that the substitution of these traditional brick kiln practices such as Bull's Trench Kiln (BTK) and Fixed Chimney Bull's Trench Kiln (FCBK) adopted in the South-Asia's IGP region with more efficient technologies such as ZigZag or Hoffman Kilns may reduce CO and PM emissions by over 60\% \cite{seay2021impact, kumar2021climate} . Despite China being the world’s largest producer of brick kilns, its manufacturing is dominated by modern technologies using Hoffman Kilns, allowing it to reduce emissions of toxic pollutants and greenhouse has emissions \cite{bashir2023investigating}. Therefore, by determining the location of these polluting assets, we can estimate pollution exposure to nearby population and demand for better regulations.\\

\noindent Previously, there have  been efforts to detect brick kilns at scale using satellite imagery in Bangladesh \cite{lee2021scalable} and India \cite{paul2022brick}. An emission inventory developed for the Indian region of IGP has declared brick kilns to be the top-most priority for assessing particulate matter emissions with Uttar Pradesh, India having the highest number of reported brick kilns \cite{ghosh2024district}. Previous works have used field surveys and remote sensing to evaluate energy consumption and enumerate brick kilns in India \cite{tibrewal2023reconciliation}. Deep Neural Networks (DNNs) have been used to demonstrate the use of high-resolution satellite imagery for the West Bengal region in India \cite{paul2022brick}; whereas, another study highlights transfer learning and self-supervised learning techniques for detecting 7477 brick kilns in five states of India with an accuracy of 81.72\% \cite{mondal2024scalable}. For Bangladesh, high-resolution satellite imagery and a deep learning backed approach have been used for the identification of brick kilns, reporting 94.2\% accuracy with 88.7\% precision \cite{lee2021scalable}. \\

\noindent In Pakistan, despite efforts to use satellite imagery and AI to detect brick kilns in certain regions of the country \cite{Tahir,nazir2023mitigatingclimatehealthimpact}, there remains a data gap in large-scale satellite monitoring of brick kilns. \\

\noindent To this end, we developed an asset-level pipeline for detecting air pollution from brick kilns, consisting of three main stages: (1) an initial detection using a Random Forest model on low-resolution Sentinel-2 satellite imagery, followed by a multi-step post-processing pipeline to reduce the high false-positive rate in the raw outputs; (2) identified points are then analyzed with high-resolution imagery from the Google Maps Static API, where a YOLOv8 object detection model differentiates between Fixed Chimney Bull's Trench Kiln (FCBK) and ZigZag kiln types, further validating the detections and enabling accurate identification in areas where Sentinel-2 data is not available; and, finally (3) we estimate the industry’s air pollution contribution by calculating emissions from each identified kiln, quantifying metrics for PM$_{10}$, PM$_{2.5}$, SO$_x$, and NO$_x$.

\section*{Methods}
We present a scalable pipeline combining machine learning and deep learning techniques for the accurate identification of brick kilns across Pakistan’s Indo-Gangetic Plain (IGP). The Pakistani portion of the IGP primarily consists of the vast Indus River plain, covering approximately 200,000 square miles (518,000 square kilometers) of fertile land \cite{burki_ziring_2024}. This region encompasses a significant part of Pakistan’s agricultural heartland, including most of Punjab province, as well as part of Sindh and Khyber Pakhtunkhwa (KP) province.\\

\noindent Although brick kilns are easily visible in satellite images, precisely identifying their locations is time-consuming and expensive. To address this issue, we utilize a combination of openly available low-resolution satellite imagery and commercial high-resolution satellite imagery that helps to reduce costs and speed up the identification process.  \\

\noindent The methodology is divided into two phases to effectively identify and classify brick kilns in satellite imagery. In the first phase, we applied a Random Forest classifier to low-resolution RGB Sentinel-2 imagery for initial pixel-wise detection followed by a post-processing step across the entire region shown in Figure \ref{fig:igp}(b). This allowed for extensive coverage but was hindered by the similarity in color profiles between brick kilns and surrounding areas, leading to occasional misclassifications. To overcome this, we integrated a YOLOv8 object detection model, using high-resolution imagery from the Google Maps Static API to differentiate between FCBK and ZigZag kilns and validate points identified by the Random Forest model. This high-resolution pipeline was also utilized in regions without Sentinel-2 data. By primarily relying on open-source low-resolution imagery and selectively using high-resolution imagery when necessary, we created a comprehensive dataset of brick kiln locations across the study area, addressing the existing data gap in Pakistan.

\subsection*{Phase One: Random Forest for Pixel-Wise Identification}
In this phase, we leveraged Random Forest to perform an initial, pixel-wise classification of brick kilns using Sentinel-2 imagery followed by a dedicated post-processing pipeline to refine detections, reduce false positives, and enhance the accuracy of the identified kiln locations.

\subsubsection*{Data Collection and Annotation for Random Forest}
The data collection process focused on acquiring satellite imagery for the study area in the Pakistani IGP region. We utilized Sentinel-2 imagery from July 1, 2023, to July 15, 2023, a period when brick kilns are typically active in Pakistan. Specifically, we used Sentinel-2 MSI (MultiSpectral Instrument) Level-2A imagery, which provides surface reflectance data for precise detection and analysis \cite{sentinel2}. 
The dataset included the RGB bands of the Sentinel-2 imagery and served as our primary dataset for brick kiln detection across most of the study area.
To ensure high-quality imagery, we applied a cloud cover removal process and downloaded only images with less than 2\% cloud cover. The cloud masking process involves identifying and removing pixels affected by clouds and cirrus from the imagery. This is achieved by evaluating quality assessment flags that indicate the presence of these atmospheric conditions. By applying this mask, we ensure that only clear and cloud-free data is utilized.
The Pakistani IGP region as shown in Figure \ref{fig:igp}b), was divided into measuring 100 x 100 km² grids, totaling approximately 60 grids. To effectively train the models, we annotated satellite images of each grid tile with different land cover classes. This process involved manually labeling the images to categorize different land cover types, such as urban areas, vegetation, water bodies, bare land, and brick kilns as the main target feature. Careful annotation was crucial to avoid confusion in the model training process and to ensure the accurate classification of diverse land cover types within the area of interest. The visual appearances of these land cover types are illustrated in Figure \ref{figure_classification} and their semantic characteristics are delineated in Table \ref{Semantic}. By annotating all 60 grids across the entire region rather than focusing on a smaller subset, this approach is estimated to have reduced false positives by at least threefold. We categorized the annotations into ten classes as shown in Table \ref{number_of_pixels}.

\subsubsection*{Training Setup and Pixel-Wise Classification}
For our classification task, we employed a Random Forest classifier, a robust ensemble learning method known for its ability to handle diverse datasets. This model was selected due to its proven efficacy in managing high-dimensional data, such as the Sentinel-2 imagery, and its capability to accommodate both continuous and categorical variables \cite{sentinel_efficacy}. The dataset was divided into 80\% training and 20\% testing, ensuring that the model was trained on a substantial portion while reserving enough data for evaluation. The Random Forest model was trained with 500 decision trees (estimators), each constructed using a random subset of the training data and features to mitigate over-fitting and enhance the model’s generalisability. At each split, a maximum of ten features were considered, with a maximum tree depth of 50, a minimum of two samples per split, and a minimum of one sample per leaf. This configuration promoted efficient feature selection and optimized model performance on the test data. \\

\noindent The model achieved a high recall of 97\% and a precision of 72\% on the test dataset for detecting brick kilns. A recall rate of 97\% indicates that the model successfully identified 97\% of the actual brick kiln pixels in the dataset, meaning that it captured nearly all instances of the brick kilns, with very few false negatives. However, the precision rate of 72\% shows that, among the instances labeled as brick kilns by the model, only 72\% were actually brick kilns, indicating a considerable number of false positives. In this context, a false positive occurs when the model incorrectly labels a non-brick kiln feature as a brick kiln. High recall with low precision often implies that while the model is very sensitive in detecting potential brick kilns, it lacks specificity, and mistakenly identifies other structures or features with similar visual characteristics as brick kilns. These false positives can lead to overestimations in the count or area of brick kilns, which would skew the analysis and limit the practical utility of the results. To address this imbalance and improve the accuracy of our brick kiln detection, we recognized the need for a post-processing pipeline.

\subsubsection*{Sentinel-2 Grid Selection For Inferencing}
Initially, we used 5x5 km grids to cover large areas, but this grid size proved inadequate for detecting smaller, dispersed kilns due to Sentinel-2’s 10m resolution. As shown in Supplementary Figure 1, where (a) represents a 5x5 km grid and (b) represents a 1x1 km grid, the larger grid size makes it challenging to visually identify individual brick kilns. The 5x5 km grid limited the effectiveness of post-processing steps, resulting in reduced detection accuracy. Switching to 1x1 km grids allowed for more focused analysis, better aligned with Sentinel-2's resolution, and improved the precision of our post-processing steps. This adjustment reduced false positives and allowed us to capture smaller kilns more accurately. Consequently, the 1x1 km grid proved to be the optimal choice for accurate brick kiln identification across the study area.

\subsubsection*{Post Processing and Geolocating}
After training the model, we applied the Random Forest classifier to Sentinel-2 imagery across the entire IGP region, where each Sentinel-2 image covers an area of 1x1 \text{km}\textsuperscript{2}.
Following the initial pixel-wise classification using Random Forest across the entire IGP region as illustrated in Figure \ref{overview_pipeline}(a), we applied a post-processing step to enhance the accuracy of brick kiln detection. This step was crucial for reducing noise and improving the coherence of detected regions. The post-processing pipeline, shown in Figure \ref{overview_pipeline}(b), follows a structured sequence of steps. The process begins by generating a binary mask from the Random Forest model's pixel-wise classification results, which identifies areas that are likely to contain brick kilns. This results in a new image consisting of two classes: detected brick kilns, shown as red pixels, and the background, represented in black. Isolated pixels are removed to reduce false positives. Morphological closing is applied to eliminate noise and consolidate fragmented areas, forming coherent clusters. These red pixels are clustered based on proximity, with each cluster corresponding to a distinct kiln. The geometric center of each cluster is computed to determine precise kiln locations, converted into geographical coordinates, and redundant detections within a 20 meter radius are eliminated. Finally, the number of centers is capped at fifteen per square kilometer, as it is highly improbable for more than fifteen brick kilns to exist within such a confined area, ensuring a significant reduction in false positives.

\subsubsection*{Sentinel-2 Pipeline Results}
The machine learning pipeline was evaluated across four regions: Khyber Pakhtunkhwa (KP), northern Punjab, southern Punjab, and Sindh. The entire IGP region comprises 20,873 data points, with 11,536 points from southern Punjab and Sindh, and 9,337 points from Northern Punjab and KP. The model performed particularly well in Northern Punjab and KP, as the detected points closely matched the expected number of brick kilns in these areas. However, in Southern Punjab and Sindh, the model produced a higher rate of false positives. Despite this, the post-processing pipeline significantly improved results, reducing false positives by twenty-fold. Without this pipeline, we estimated that over 400,000 points would have been generated, making cross-verification prohibitively costly. There were also areas where Sentinel-2 data was unavailable and the model could not detect any brick kilns in those regions.

\subsection*{Advancing to YOLO: Building on Random Forest Findings}

In the random forest phase, detecting brick kilns in the Sindh and Southern Punjab regions presented a challenge—kilns and surrounding areas exhibiting red hue. This resemblance led to high false positive rates. Since the random forest algorithm analyzes each pixel independently, it loses important spatial context, making it difficult to distinguish between neighboring objects. Additionally, though fewer, false positives also occurred in Northern Punjab and KP. Given these challenges and the need to classify brick kilns into FCBK and ZigZag types, we transitioned to YOLO, which improves spatial awareness by dividing the image into grids, predicting bounding boxes, and assigning class probabilities, resulting in more accurate object detection. YOLO’s deep learning architecture autonomously learns relevant features during the training process, eliminating the need for manual feature selection required in methods such as Random Forest. Our kiln type identification pipeline was ran on two types of regions: (a) 20,873 locations identified by our Random Forest model and (b) the region where Sentinel-2 imagery was unavailable (Supplement Figure S2).

\subsection*{Phase Two: YOLO for High-Resolution Imagery}
In this phase of the pipeline, we used high-resolution satellite imagery and YOLO to enhance the detection of brick kiln points identified in low-resolution imagery by the Random Forest classifier.
\subsubsection*{Data Collection and Annotation for YOLO}
We extracted high-resolution imagery through the Google Maps Static API, using a zoom level of 17 and a scale of 2, which produced images with dimensions of 1280x1280 pixels. The scale factor effectively doubled the resolution from the default 640x640 pixels, providing a greater level of detail. For these high-resolution images, the annotation process was straightforward. We annotated approximately 375 FCBK brick kilns and 295 ZigZag brick kilns using bounding boxes. While Oriented Bounding Boxes (OBBs) generally provide a more precise representation of an object’s shape and orientation \cite{orientationboundingboxes}, we opted for standard bounding boxes to streamline geolocation. Since our objective was to obtain only a single central coordinate per kiln, the added computational expense of OBBs was unnecessary, making regular bounding boxes a more efficient choice.

\subsubsection*{Training Setup}
The YOLOv8n \cite{Jocher_Ultralytics_YOLO_2023} model was trained using a dataset split into 80\% for training, 10\% for validation, and 10\% for testing, ensuring robust evaluation and fine-tuning throughout the process. The model was trained for 250 epochs, with a batch size of 8 and an image resolution set to 1280×1280. Early stopping was applied with patience of 100 epochs, and the model checkpoint from epoch 114 was selected due to no further performance improvement after this point. The training utilized an initial learning rate (lr0) of 0.01 with a learning rate decay factor (lrf) of 0.01. The optimizer was set to auto, with a momentum value of 0.937 and a weight decay of 0.0005. A warmup period of 3 epochs was employed, with a warmup momentum of 0.8 and a warmup bias learning rate of 0.1. Data augmentation techniques included RandAugment with a probability of erasing set to 0.4, translation at 0.1, and scaling at 0.5. The mosaic augmentation was activated with a factor of 1.0, and horizontal flipping was applied with a probability of 0.5. Other transformations, such as rotation (degrees) and shear, were kept at 0.0. Non-maximum suppression (NMS) was configured with an IoU threshold of 0.7, and the maximum number of detections per image was capped at 10. The model used overlap masks with a ratio of 4 and was trained with 8 workers. Automatic Mixed Precision (AMP) was enabled to optimize computational efficiency. \\

\noindent The YOLO object detection model was trained using High-Resolution satellite imagery. The training yielded strong results, with a mean Average Precision (mAP) at 50\% Intersection over Union (IoU) reaching 95\%, and mAP at 95\% IoU reaching 52\%. The normalized confusion matrix in Figure \ref{cf} illustrates the model's classification performance across the three classes on the test dataset: FCBK, ZigZag, and background. A high-level view of obtaining class and bounding box predictions is shown in Figure \ref{overview_pipeline}(c).

\subsubsection*{Inference}
For the YOLO pipeline, inferencing was conducted on two key areas: the 20,873 points identified by our Random Forest model and regions where Sentinel-2 data was unavailable. Among the 20,873 identified points, some were located in close proximity, raising the risk of overlapping brick kiln detections. To mitigate this, we used a zoom level of 17, where each image covers an area of 0.45 km². We grouped coordinates within a 0.45 km² radius, downloading a single image per group to minimize redundancy and optimize data management. This grouping approach effectively reduced double counting and minimized the volume of imagery required, while addressing the high false positive rate observed in this region. Additionally, if a kiln appeared split across two contiguous images as shown in Figure \ref{half_kiln}a) and Figure \ref{half_kiln}b), only one instance was counted by removing points within a 12-meter radius of each other. To accurately calculate the geographic coordinates for each detected bounding box, we applied a method described in the Supplementary Section I, which maps bounding box centers to geographic coordinates (Equations 1-6). This approach ensures precise localization of brick kilns within the satellite imagery. \\

\noindent For areas lacking Sentinel-2 coverage, high-resolution imagery was obtained for the entire region to ensure comprehensive detection of brick kilns. After the completion of the second phase of the entire pipeline, we detected more than 11,000 brick kilns in the region.\\

\section*{Data Records}
The dataset produced for this study captures critical attributes related to brick kiln air pollution-related emissions and their potential risk to the nearby population. The data records consist of geolocation information including coordinates, country, and the district, see Figure \ref{type}(a). The database also includes a detailed profile of brick kiln sites, emission estimates, and their proximity to sensitive areas such as schools, hospitals, and the populated zones (within a 1 km radius) as shown in the Figure \ref{type}(c). All the amenities and the population within close proximity to brick kilns are at a heightened risk of exposure to air pollution, which can affect the respiratory health of the population and overall well-being.
For emission estimates, the data record includes pollutants such as PM\textsubscript{10}, PM\textsubscript{2.5}, SO\textsubscript{x}, and NO\textsubscript{x} for each brick kiln site. These emissions were calculated using a bottom-up approach based on production data and standardised emission factors, accounting for operational days and the types of fuel used during kiln operations. See Supplementary Section II for more details. It is expressed in kilograms per day, providing insight into potential daily pollutant emissions for each site.
These measures are helpful in the assessment of the number of people potentially impacted by emissions and can be crucial for developing targeted regulatory or mitigation strategies. Additionally, each kiln in the dataset is categorized by its operational type: FCBK and ZigZag as shown in the Figure \ref{type}(b). This classification is essential as different kiln types have varying fuel efficiencies and emission profiles. Zigzag kilns, for example, are generally more fuel-efficient and emit fewer pollutants than traditional kilns. This attribute enables comparative analysis across kiln types to assess the environmental impact and will assist regulatory bodies in identifying kilns that may benefit from technological upgrades.

\section*{Technical Validation}
The annotations for both our Random Forest and YOLO models were performed by experts familiar with satellite imagery and brick kiln detection. To ensure consistency and reduce annotation errors, the annotations were cross-validated by different team members. This process helped establish a strong ground truth dataset for training and validating our models. Our validation process was designed in two parts: before running inference over the entire region and after the full inference was completed. Prior to the large-scale inference, we evaluated the performance of our models using precision, recall, and F1 scores across different regions. This pre-inference validation was critical for assessing whether the model was adequately trained and performing as expected. As the Table \ref{table:precision_recall} indicates, we observed a high number of false positives from the Random Forest Classifier which necessitated further refinement through the post-processing pipeline outlined in the paper. The final results, after running both the low-resolution (Sentinel-2 and Random Forest) and high-resolution (YOLOv8) pipelines, were manually verified by experts to ensure the accuracy of detected kiln locations.\\

\noindent As outlined in Table \ref{DetectionRates}, the trained YOLO model successfully detected approximately 11,277 brick kilns. Of these, 6,706 were located in Northern Punjab and Khyber Pakhtunkhwa (KP), while 4,271 were detected in Sindh and Southern Punjab. These findings align with our initial hypothesis regarding the performance of the Random Forest classifier. In Northern Punjab and KP, where the Random Forest Classifier identified 9,337 points, 6,706 were confirmed to be brick kilns, indicating strong performance in these regions. Conversely, the model showed moderate performance in Sindh and Southern Punjab, with 4,271 brick kilns detected out of 11,536 points. Due to the effectiveness of the initial post-processing pipeline explained earlier, we downloaded a total of 60,000 high-resolution images for analysis including the regions where Sentinel-2 data was unavailable reducing the costs associated with the high-resolution imagery. \\

\noindent For emissions estimates, we validated by comparing our bottom-up approach results with previous studies on brick kiln emissions in similar regions \cite{ABBAS}. The proximity of schools, hospitals, and populated areas within 1 km of each kiln was assessed using the kiln points identified in our findings, combined with OpenStreetMap (OSM) data and (GIS) techniques. \\

\noindent A few limitations to this pipeline are delineated in Supplementary Section III. \\

\section*{Usage Notes}
The code for the project is available on GitHub, with detailed instructions for setup and usage. For the low-resolution pipeline, we used Google Earth Engine’s Python-based API to download GeoTIFF files from Sentinel-2 imagery. Rasterio was employed for geospatial data handling, Scikit-learn library in Python was used to train the Random Forest Classifier, and Python OpenCV was used for post-processing steps such as noise removal and clustering of detected brick kilns. For the high-resolution pipeline, we used Ultralytics' Python API to handle the training and inferencing of the YOLOv8 model. High-resolution imagery from Google Maps Static API was processed to improve detection accuracy. The dataset is provided as a CSV file with three columns: the first indicating the kiln type (0 for FCBK and 1 for Zigzag), followed by the latitude and longitude of each detected kiln. This format allows for easy integration with GIS software or further analysis. \\

\noindent This dataset is provided in three formats: CSV file containing multiple attributes for each brick kiln, shapefile (.shp), and geojson(.geojson) for spatial analysis, which includes location points and associated attributes such as emission estimates, kiln type, and proximity to sensitive sites. Emission estimates are presented in units of grams per kilogram (g/kg) of bricks produced, with the spatial data using the WGS84 coordinate reference system (CRS). Users can access and download these files through our \href{https://zenodo.org/records/14038648}{Zenodo Data Records} for integration into GIS and data analysis software, such as QGIS, ArcGIS, or R/Python. For further details on attribute descriptions, units, and abbreviations, please refer to the supplementary document S1. 

\section*{Code availability}
The code including model weights for this paper are open-source and available at \href{https://zenodo.org/records/14038648}{Zenodo Data Records}.

\section*{Recommended Use}

Researchers can utilize the \textit{csv}, \textit{shapefile}, or \textit{geojson} as they prefer, to analyze spatial distributions, pollutant estimates, and proximity risks. Users can conduct spatial queries, overlay additional environmental layers, and perform risk assessment modeling. For accurate and up-to-date demographic data, users may cross-reference with local census or updated OSM data, especially for studies involving dynamic population or infrastructure changes.

\section*{Acknowledgements} 
We would like to thank Smith School of Enterprise and the Environment and Amazon Web Services (AWS) for funding and support respectively. We also appreciate AWS Cloud Credit for Research.

\section*{Author contributions statement}
M.S.A.H developed machine learning and deep learning pipelines. K.Z. worked on the remote sensing pipeline. M.S.A.H., K.Z., N.K. S.E.F., annotated the imagery. K.Z., M.S.A.H., S.E.F validated the results. M.S.A.H. and K.Z. wrote the first draft. H.A.S. and N.K commented on the paper. H.A.S. conceptualised the need for a brick kiln database for Pakistan and supervised the project. All authors reviewed the paper..

\section*{Competing interests} 
No competing interests

\bibliographystyle{plain}
\bibliography{main}

\section*{Figures \& Tables}

\begin{figure}[h!]
    \centering
    \includegraphics[width=1\textwidth]{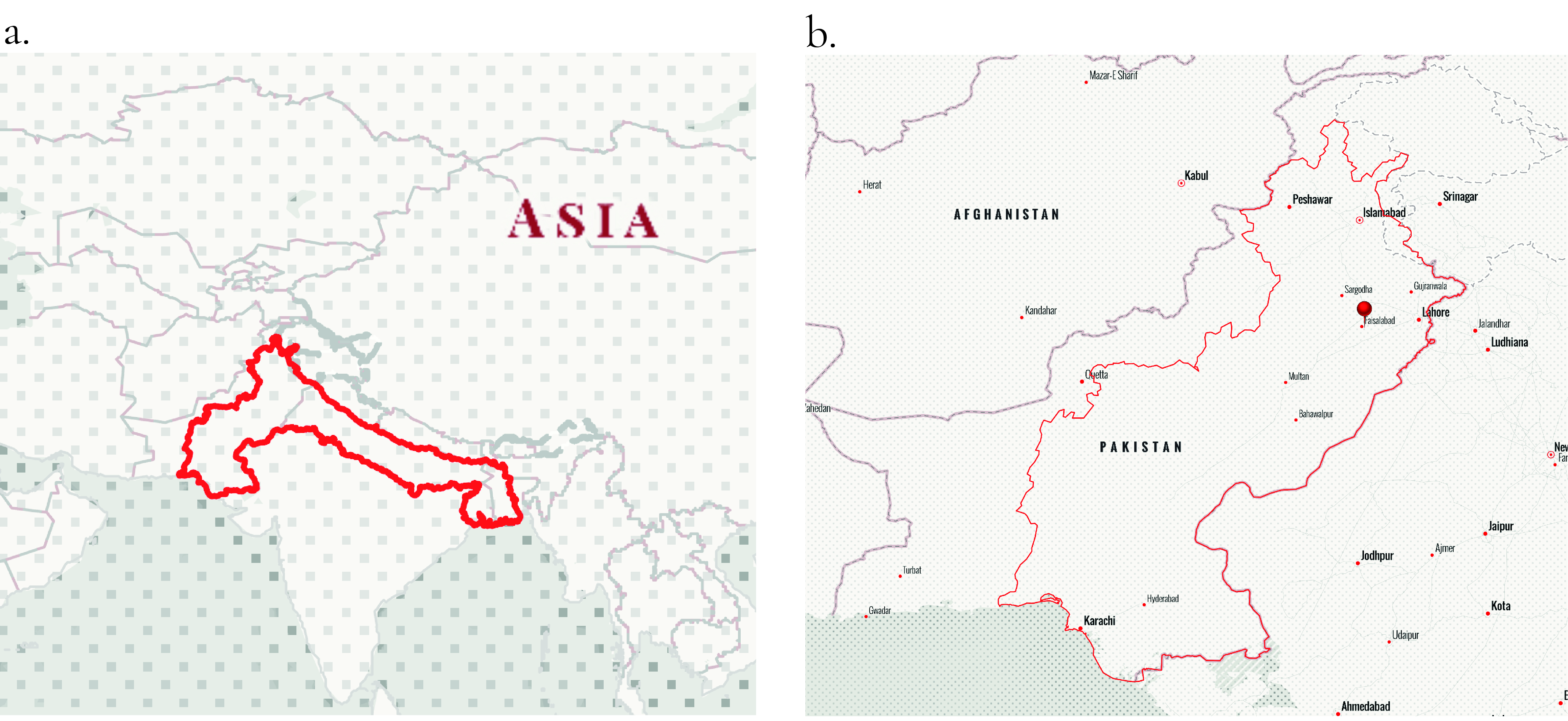} 
    \caption{Study Area Map - (a) The Indian-Gangetic Plain (IGP) region, spanning across Bangladesh, India, and Pakistan (b) Focuses on the IGP region within Pakistan, which is the specific area of interest for kiln detection in this study.}
    \label{fig:igp}
\end{figure}

\begin{figure}[h!]
    \centering
    \includegraphics[width=0.8\textwidth]{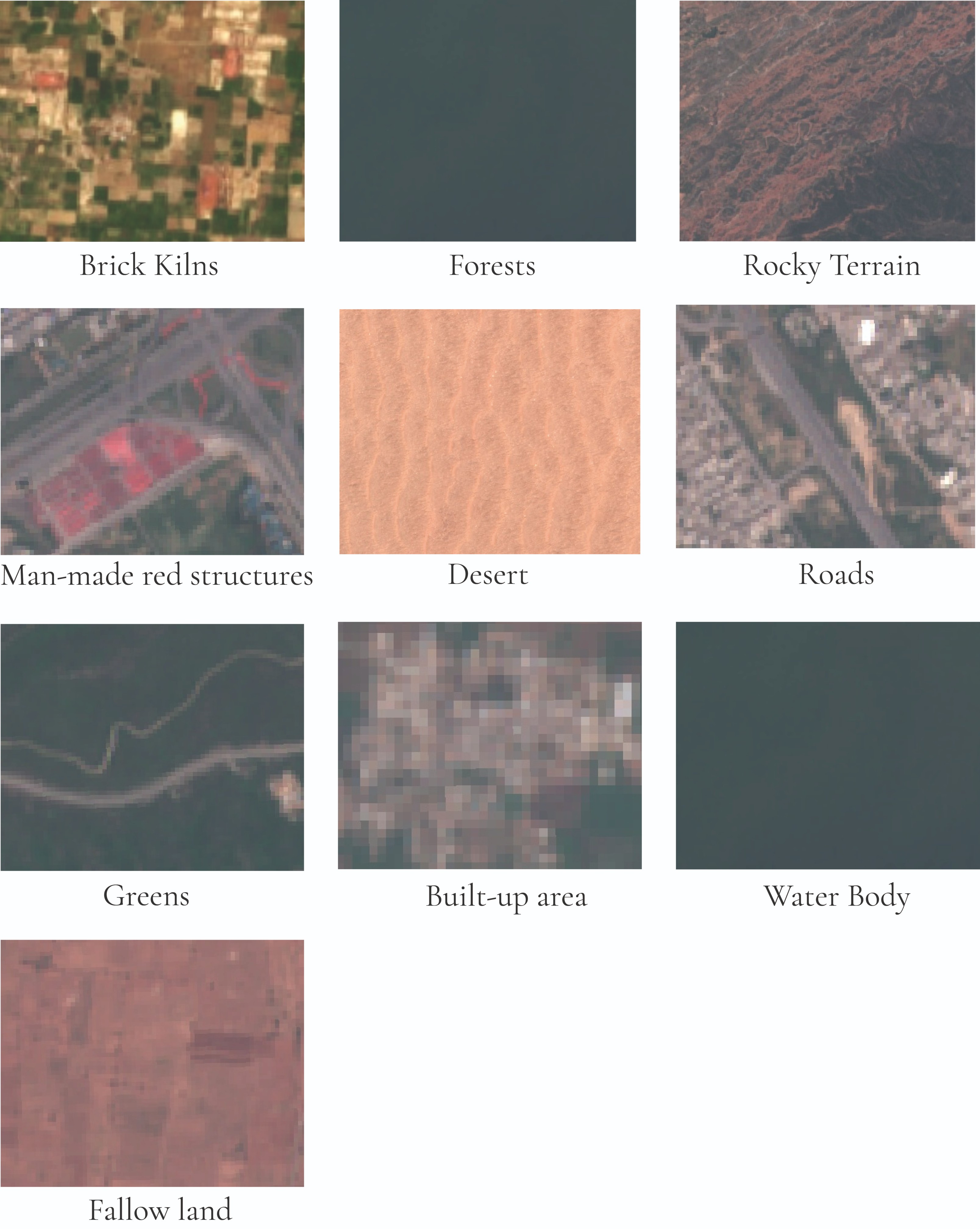} 
    \caption{Image tiles illustrating different classes labelled for training our Random Forest Classifier}
    \label{figure_classification}
\end{figure}

\begin{figure}[h!]
    \centering
    \includegraphics[width=0.7\textwidth]{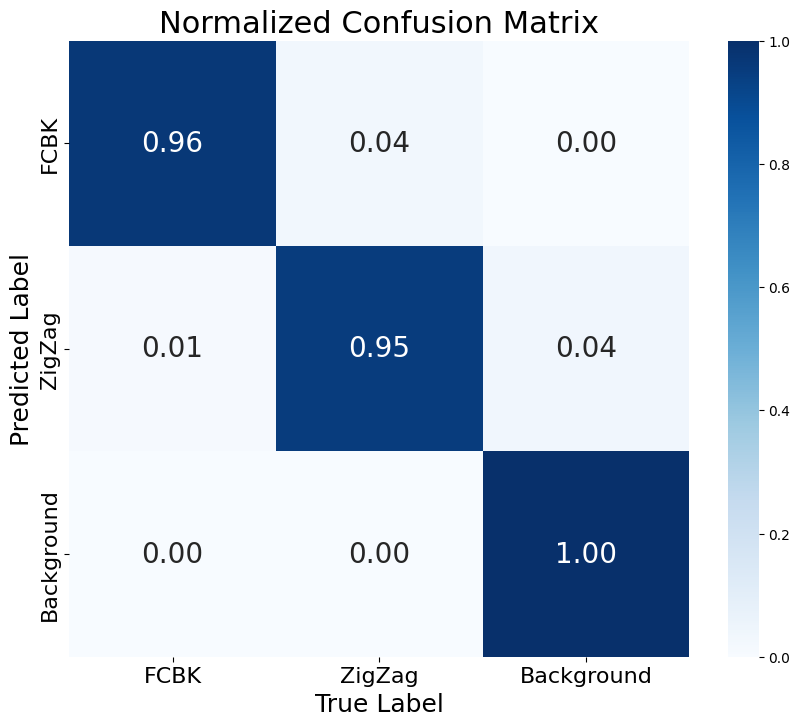} 
    \caption{Normalized Confusion Matrix for YOLOv8n.}
    \label{cf}
\end{figure}

\begin{figure}[h!]
    \centering
    \includegraphics[width=1\textwidth]{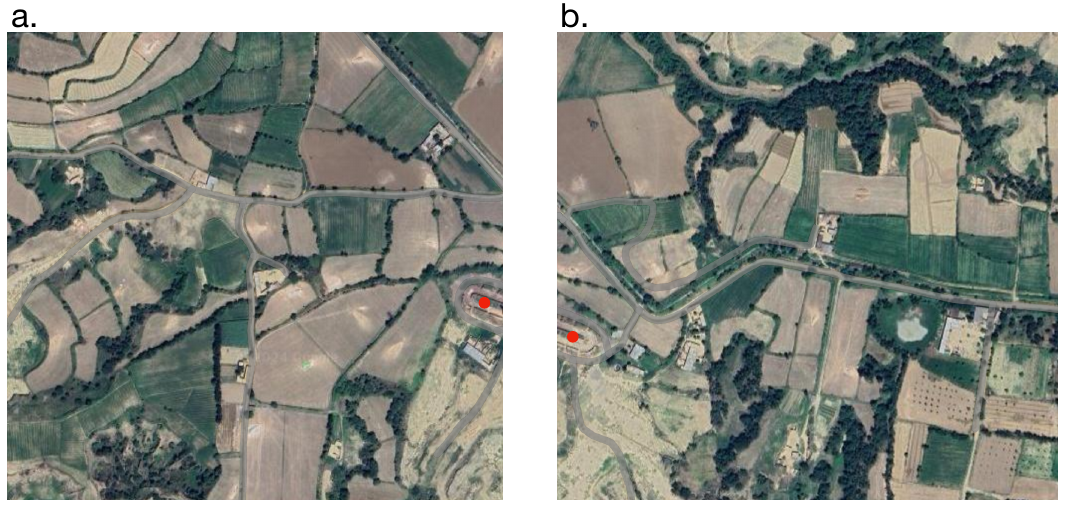} 
    \caption{a) and (b) represent two adjacent image tiles where the same kiln is partially visible
in both.}
    \label{half_kiln}
\end{figure}

\begin{figure}[h!]
    \centering
    \includegraphics[width=1\textwidth]{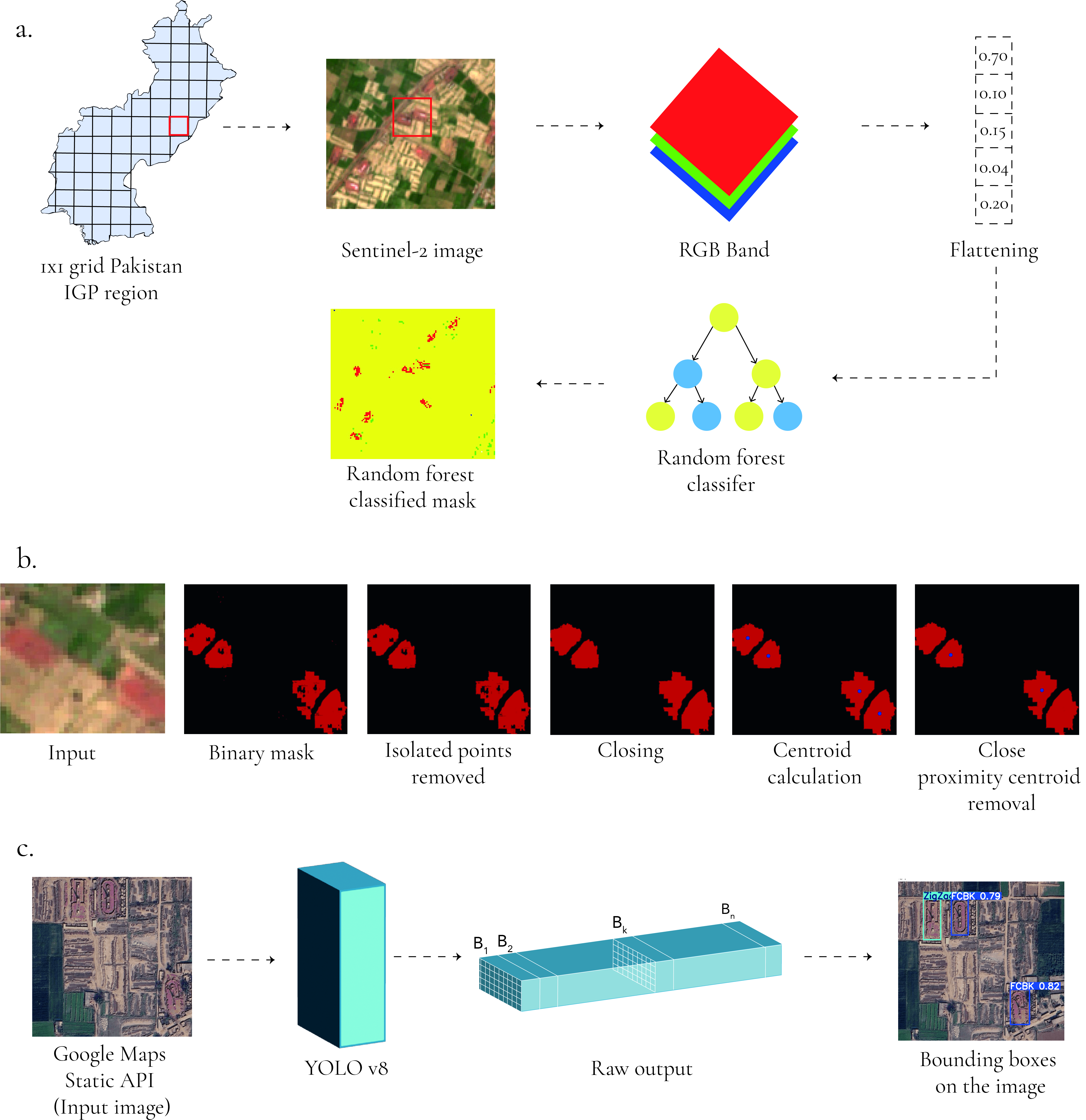} 
    \caption{a) The figure outlines the process of brick kiln detection in the Indus-Gangetic Plain using Sentinel-2 imagery. A 1x1 grid of the region is extracted, RGB bands are flattened into feature vectors, and a Random Forest classifier generates a mask identifying brick kiln locations. b) The figure shows the visual representation of the steps involved in post-processing pipeline to accurately geolocate brick kilns in the image. c) The figure illustrates the process of using the YOLOv8 model to detect brick kilns from Google Maps imagery, resulting in bounding boxes on the input image.}
    \label{overview_pipeline}
\end{figure}

\begin{figure}[h!]
    \centering
    \includegraphics[width=1\textwidth]{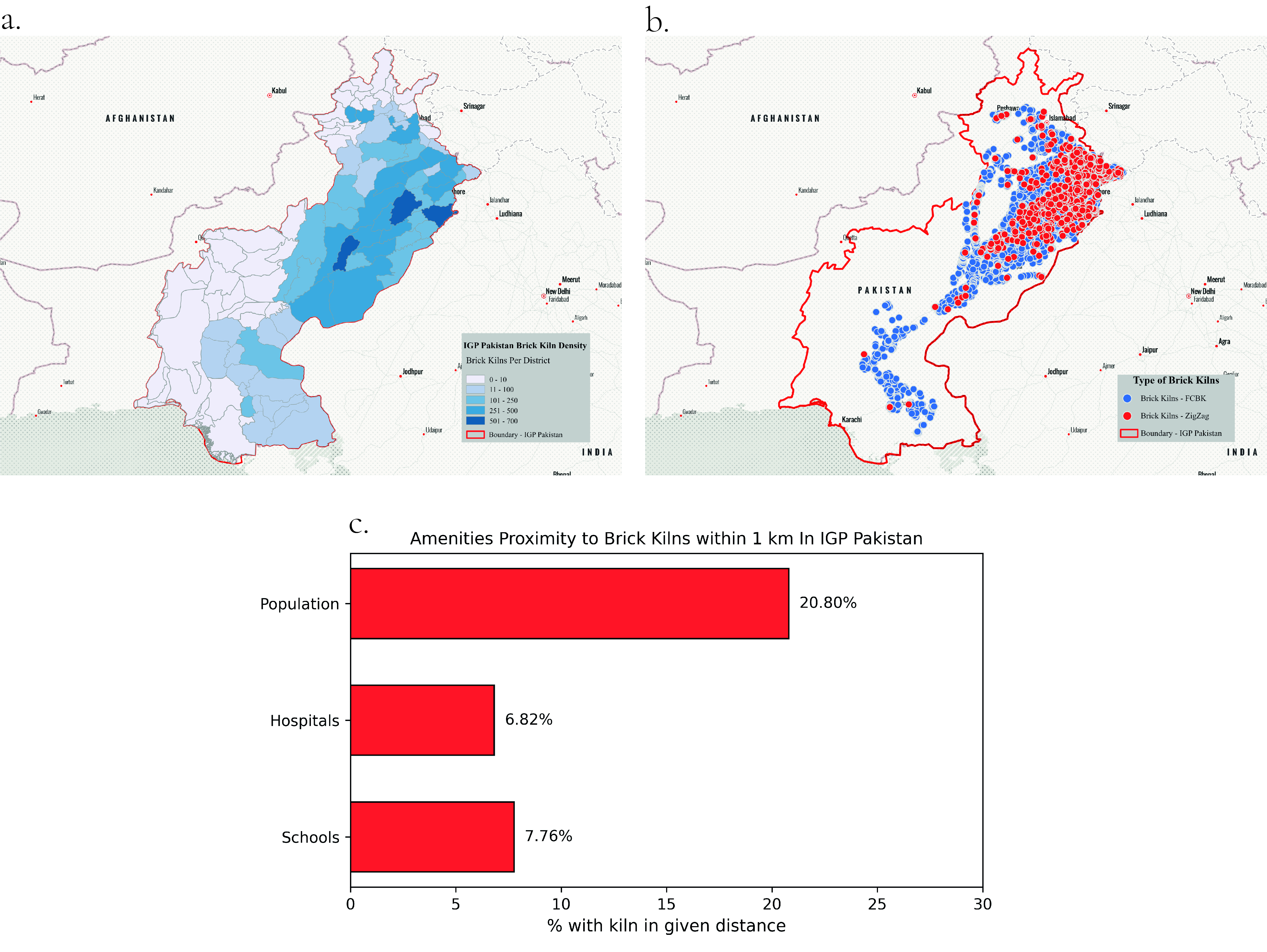} 
    \caption{Overview of brick kiln distribution, types, and proximity to sensitive areas within IGP-Pakistan. 
    (a) Distribution of brick kiln density across IGP-Pakistan, highlighting areas with high kiln concentrations and potential pollution hotspots. 
    (b) Classification of kiln types (FCBK and zigzag) showing regional differences in kiln technology.
    (c) Percentage of schools, hospitals, and population density within 1 km of kilns, indicating exposure risks for nearby communities.}
    \label{type}
\end{figure}

\newcolumntype{L}[1]{>{\raggedright\arraybackslash}m{#1}}

\renewcommand{\arraystretch}{1.5}

\begin{longtable}{|L{1.5cm}|L{2.5cm}|L{2.5cm}|L{6.5cm}|}
\hline
\textbf{T0} & \textbf{T1} & \textbf{T2} & \textbf{Semantic Characteristic} \\
\hline
\endfirsthead

\hline
\textbf{T0} & \textbf{T1} & \textbf{T2} & \textbf{Semantic Characteristic} \\
\hline
\endhead

\endfoot

\endlastfoot

\multirow{10}{*}{\makecell[l]{Land\\ Cover}} & 
\multirow{2}{*}{\makecell[l]{Vegetated\\ Areas}} & 
Green Areas & Areas covered with grass and maintained vegetation with tones of green \\
& & Forests & Large areas with continuous dense, dark green patches with dominance of trees \\
\cline{2-4}

& \multirow{3}{*}{\makecell[l]{Non-\\ Vegetated Areas}} & 
Fallow Lands & Light brown, reddish tones with large, open rectangular patches with visible soil and sparse or no vegetation \\
& & Desert & Uniform land with a sandy texture, little vegetation, brown or beige tones, and ripple like patterns expanding large areas with little variation in landscape  \\
& & Rocky Terrain & Rough, jagged patterns with reddish brown tones and uneven textures with pronounced vertical features \\
\cline{2-4}

& \multicolumn{2}{|l|}{Water Bodies} & Rivers, lakes, ponds, and other standing or flowing bodies of water \\
\cline{2-4}

& \multirow{3}{*}{\makecell[l]{Urbanized\\ Structures}} & 
Urban Areas & Pixelated areas with mixtures of colors appear as dense and complex patches of structures \\
& & Redroof Structures & Red human made structures with rectangular or geometric patterns with visible boundaries \\
& & Roads & Long, straight, or curved linear features typically in gray tones sometimes with intersections \\
\cline{2-4}

& \multirow{1}{*}{Industrial Use} & Brick Kilns & Ovular or rectangular structures surrounded either by vegetation or barren land with a distinguishable reddish-brown tone. \\
\cline{3-4}

\hline

\caption{Categorisation of Land Cover Types with Semantic Characteristics for Each Sub-Class} \label{Semantic}
\end{longtable}

\begin{table}[h]
\centering
\renewcommand{\arraystretch}{1.8} 
\begin{tabular}{p{3cm} p{4cm} p{4cm} p{3cm} }
\hline
\textbf{Region} & \textbf{Total Points Identified by RF} & \textbf{Brick Kilns Detected by YOLO} & \textbf{Percentage Detected (\%)} \\ \hline
Northern Punjab and KP & 9,337 & 6,706 & 71.8\% \\ \hline
Sindh and Southern Punjab & 11,536 & 4,271 & 37.0\% \\ \hline
Regions without Sentinel-2 Imagery (Google Maps Static API only) & - & 301 & - \\ \hline
\textbf{Total} & \textbf{20,873} & \textbf{11,277} & \textbf{54.0\%} \\ \hline

\end{tabular}
\caption{Performance of Random Forest Model in Identifying Brick Kiln Points Across Different Regions and Subsequent YOLO Detection Results}
\label{DetectionRates}
\end{table}

\begin{table}[h]
\centering
\renewcommand{\arraystretch}{1.8} 
\begin{tabular}{l l r}
\hline
\textbf{Index} & \textbf{Class} & \textbf{Number of Samples (Pixels)} \\ \hline
1 & Brick Kilns & 25,678 \\ 
2 & Redroof Structures & 950 \\ 
3 & Water Bodies & 6,578 \\ 
4 & Green Areas & 13,234 \\ 
5 & Forests & 4,300 \\ 
6 & Fallow Lands & 21,893 \\ 
7 & Desert & 6,918 \\ 
8 & Urban Areas & 5,435 \\ 
9 & Roads & 3,462 \\ 
10 & Rocky Terrain & 8,214 \\ \hline

\end{tabular}
\caption{Annotated Pixel Counts for Land Cover Classes in Sentinel-2 Imagery for Random Forest Classifier Training} \label{number_of_pixels}
\end{table}

\begin{table}[h]
\centering
\renewcommand{\arraystretch}{1.8} 
\begin{tabular}{l l l}
\hline
\textbf{Region}       & \textbf{Recall} & \textbf{Precision} \\ \hline
Southern Punjab       & 0.90               & 0.65            \\ 
Sindh                 & 0.88               & 0.60            \\ 
Northern Punjab       & 0.99               & 0.78            \\ 
KP                    & 0.95               & 0.74            \\ \hline
\end{tabular}
\caption{Region-wise Precision and Recall for Brick Kilns on Test Dataset Using Random Forest Classifier}
\label{table:precision_recall}
\end{table}

\end{document}


\maketitle

\section*{Introduction}
This supplementary document provides additional details related to the methodology employed in this study and the data produced in this study that complement the primary findings presented in the main manuscript. The content herein is designed to offer a comprehensive overview of the data processing steps, mathematical derivations, and extended results referenced in the primary text. This material aims to ensure transparency in the methods used, enhance the reproducibility of the study, and support further exploration by interested researchers. 

The structure of this document is as follows: Section I describes the equations and methodology for mapping bounding box centers to geographic coordinates in satellite imagery from Google Maps Static API. Section II outlines the geolocation of brick kilns in the region of interest along with the pollutant emission estimation process from brick kiln operations, complete with equations and emission factor tables. 

\section{Supplementary Section I}
\subsection{Equations Mapping Bounding Box Center to Gographical Coordinate}
To accurately determine the geographic coordinates corresponding to bounding boxes detected in satellite imagery, we employ a methodical approach involving geospatial interpolation and image analysis. This process begins with acquiring high-resolution satellite images and performing spatial calibration for bounding box localization. Initially, a satellite image for a specified region is obtained. The latitude and longitude change per pixel are computed by measuring the geographic distance between two reference points within the image and the corresponding pixel distance. Mathematically, these quantities are expressed as:

\begin{equation}
\Delta \text{Lat}_{\text{pixel}} = \frac{\Delta \text{Lat}_{\text{geo}}}{W_{\text{image}}}
\end{equation}
\begin{equation}
\Delta \text{Lon}_{\text{pixel}} = \frac{\Delta \text{Lon}_{\text{geo}}}{W_{\text{image}}}
\end{equation}

where \(\Delta \text{Lat}_{\text{geo}}\) and \(\Delta \text{Lon}_{\text{geo}}\) represent the geographic changes in latitude and longitude over the width of the image, and \(W_{\text{image}}\) denotes the width of the image in pixels. For each bounding box detected within the image, the geographic coordinates of the bounding box center are calculated by determining the pixel offsets from the image center, which is defined as \((W_{\text{image}}/2, H_{\text{image}}/2)\), assuming a square image. The pixel offsets in geographic coordinates are computed using:

\begin{equation}
\Delta \text{Lat}_{\text{bbox}} = (\text{C}_{y} - \text{C}_{\text{center}}) \times \Delta \text{Lat}_{\text{pixel}}
\end{equation}
\begin{equation}
\Delta \text{Lon}_{\text{bbox}} = (\text{C}_{x} - \text{C}_{\text{center}}) \times \Delta \text{Lon}_{\text{pixel}}
\end{equation}

where \(\text{C}_{x}\) and \(\text{C}_{y}\) denote the pixel coordinates of the bounding box center, and \(\text{C}_{\text{center}}\) represents the center coordinate of the image.

The final geographic coordinates of the bounding box center are obtained by adding these offsets to the initial geographic coordinates of the image center:

\begin{equation}
\text{Lat}_{\text{bbox}} = \text{Lat}_{\text{center}} + \Delta \text{Lat}_{\text{bbox}}
\end{equation}
\begin{equation}
\text{Lon}_{\text{bbox}} = \text{Lon}_{\text{center}} + \Delta \text{Lon}_{\text{bbox}}
\end{equation}

where \(\text{Lat}_{\text{center}}\) and \(\text{Lon}_{\text{center}}\) are the geographic coordinates of the image center. Since images from the Google Static Maps API are geometrically flattened, the linear relationship between pixel offsets and geographic coordinates remains consistent across the image and the AOI.
\section{Supplementary Section II}
\subsubsection{Pollutant Emissions}
Assuming coal is the primary fuel used in brick kiln operations, emissions for each pollutant are calculated using established emission factors (in g/kg). These factors represent the quantity of a specific pollutant emitted in grams per kilogram of brick produced. The pollutants considered in this analysis, along with their respective emission factors, daily emissions (kg/day), and seasonal emissions (kg/year), are summarized in Table \ref{tab:emissions}.

\subsubsection*{Main Formulas}

Let \( E_i \) represent the emission factor for pollutant \( i \) in g/kg. The daily emissions \( D_i \) for each pollutant are calculated as:

\[
D_i = E_i \times \text{Total Daily Brick Weight (per kiln)}
\]

The seasonal emissions \( S_i \) for each pollutant are calculated as:

\[
S_i = D_i \times 215
\]
\subsubsection{Seasonal Brick Production}
The first step in the emission estimation process involves calculating the daily brick production per kiln, which is essential for quantifying emission levels. Since the kilns in the target region represent about 65\% of the total kilns, the seasonal brick production for this region is adjusted accordingly:

\[
\text{Total Seasonal Brick Production (65\%)} = 0.65 \times 45 \, \text{billion bricks} = 29.25 \, \text{billion bricks}
\]

Given there are 11,277 kilns in the study area, the per-day production per kiln is calculated as follows:

\[
\text{Daily Production per Kiln} = \frac{29.25 \, \text{billion bricks}}{11,277 \times 215} \approx 12,068 \, \text{bricks/day}
\]

Each brick weighs approximately 3 kg, so the total daily brick weight per kiln is:

\[
\text{Daily Brick Weight per Kiln} = 12,068 \times 3 = 36,204 \, \text{kg/day per kiln}
\]

Due to the kiln operating patterns in Sindh and Punjab, kilns do not operate continuously throughout the year. They typically cease operations during the monsoon season (July to September) and smog season (December to January). Excluding these periods results in approximately 215 operational days per year:

\[
\text{Working Days/Year (adjusted)} = 365 - 150 = 215 \, \text{days/year}
\]

\begin{table}[h!]
\centering
\begin{tabularx}{\textwidth}{|X|X|X|X|}
\hline
\textbf{Pollutant} & \textbf{Emission Factor (g/kg)} & \textbf{Daily Emissions (kg/day)} & \textbf{Seasonal Emissions (kg/year)} \\
\hline
PM\textsubscript{10} & 9.7  & 351.18  & 75,503.70  \\
PM\textsubscript{2.5} & 6.8  & 246.19  & 52,235.35  \\
SO\textsubscript{x}   & 4.6  & 166.54  & 35,810.10  \\
NO\textsubscript{x}   & 4.7  & 170.16  & 36,581.40  \\
\hline
\end{tabularx}
\caption{Emission factors, daily emissions, and seasonal emissions per kiln, assuming 215 working days per year.}
\label{tab:emissions}
\end{table}

To calculate the daily and seasonal emissions for each pollutant, the total daily brick weight per kiln is multiplied by the respective emission factor. For instance, PM\textsubscript{10} emissions are estimated at 351.18 kg/day, resulting in seasonal emissions of 75,503.70 kg/year, assuming 215 operational days. This approach is applied across all pollutants, providing an estimate of total pollutant emissions from brick kilns in the region.

\section{Supplementary Section III}
\subsection{Limitations}
Random Forest classifier was trained solely on low-resolution RGB channels from Sentinel-2 imagery and not any other spectra. After classification, we applied a post-processing step to remove outliers, identify object centroids through pixel clustering, and used Google Maps Static API to extract high-resolution imagery for YOLO-based object detection. While effective, this process is computationally intensive and time-consuming. Additionally, some Sentinel-2 imagery was not available on Google Earth Engine’s API, limiting coverage in certain regions (see Supplementary Figure S2). Incorporating additional Sentinel bands could potentially improve accuracy but would increase the complexity and cost of the pipeline. Additionally, while the study aimed to reduce false positives, the verification of false negatives—undetected brick kilns—was not studied. \\

\noindent The object detection model cannot differentiate between operational and non-operational kilns when the kiln structure remains intact. It is currently limited to distinguishing between kiln types such as Zigzag and FCBK. Future research could integrate satellite data on heat or emissions to improve detection of operational kilns.\\
 
\noindent Moreover, analysis of PM$_{2.5}$ emissions does not account for daily changes in wind direction, which may affect the exposure levels around kilns. This omission could lead to an incomplete understanding of how emissions disperse and impact surrounding areas.\\

\noindent Additionally, emission factor assumptions, derived from existing literature, may not fully represent local fuel compositions or kiln operations, leading to potential discrepancies in emission estimates. Seasonal operation estimates assume 215 operational days, excluding shutdown during monsoon and smog seasons, but actual kiln activity may vary. Similarly, proximity analysis, limited to a 1 km radius, may not capture pollutant dispersion influenced by weather and topography, and variations in population vulnerability are not accounted. Kiln type classifications, based on archival satellite imagery, may overlook recent upgrades, potentially affecting emission estimates across kiln types. Limited ground-truth data restricts the scope of validation, and the absence of real-time air quality monitoring near these kilns reduces precision, particularly for pollutants with temporal variability. However, given the scope of this study to present a more simplistic but immediate emission estimates based on the bricks identified. Undoubtedly, there is a need for more comprehensive approach to integrate all the climatic and demographic information.

\section{Supplementary Figures}
\begin{figure}[h!]
    \centering
    \includegraphics[width=0.75\textwidth]{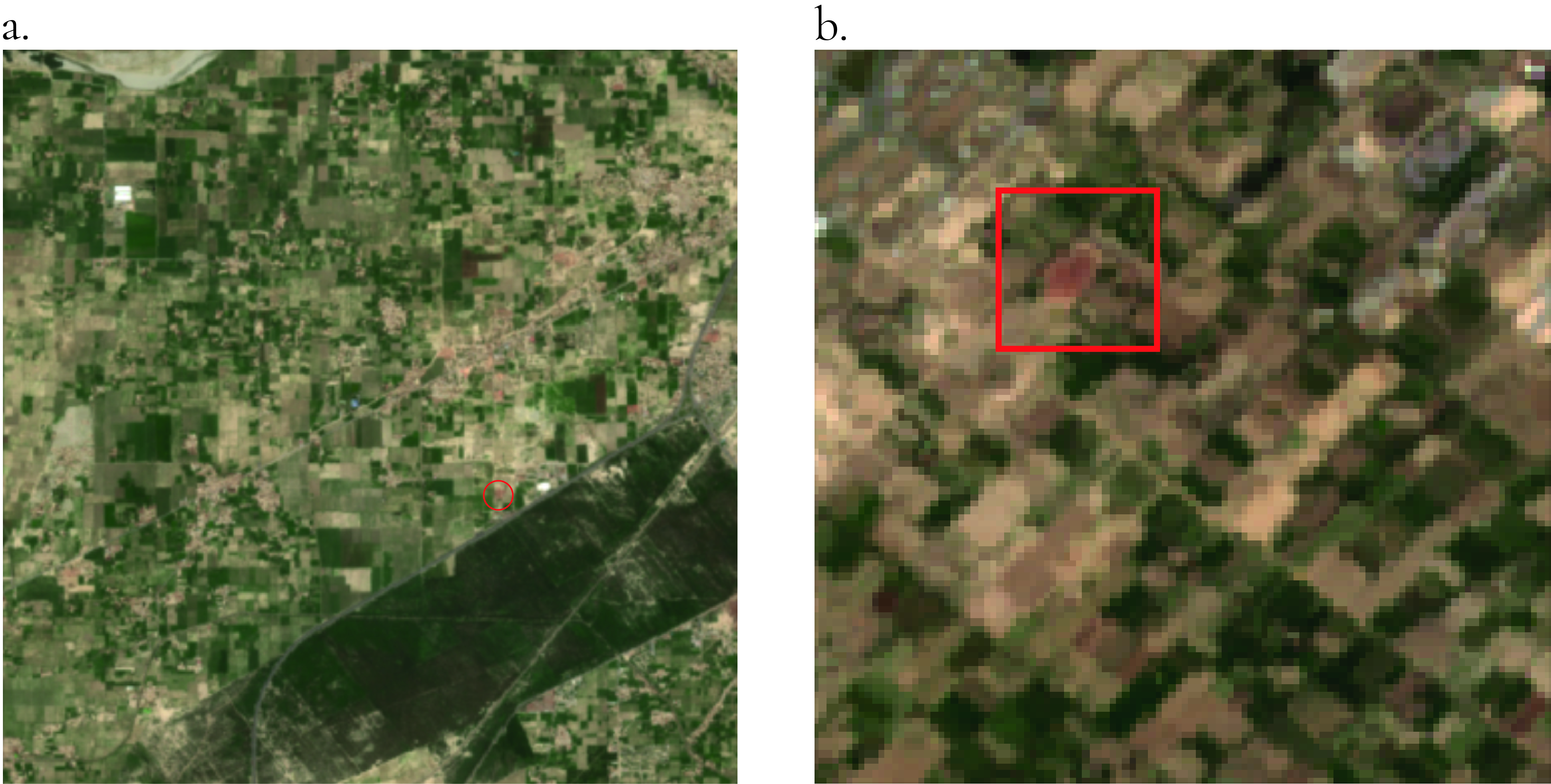}
    \caption{(a) Brick kilns highlighted within a 5x5 km grid. (b) Brick kilns highlighted within a 1x1 km grid.}
    \label{fig:grids}
\end{figure}

\begin{figure}[h!]
    \centering
    \includegraphics[width=0.75\textwidth]{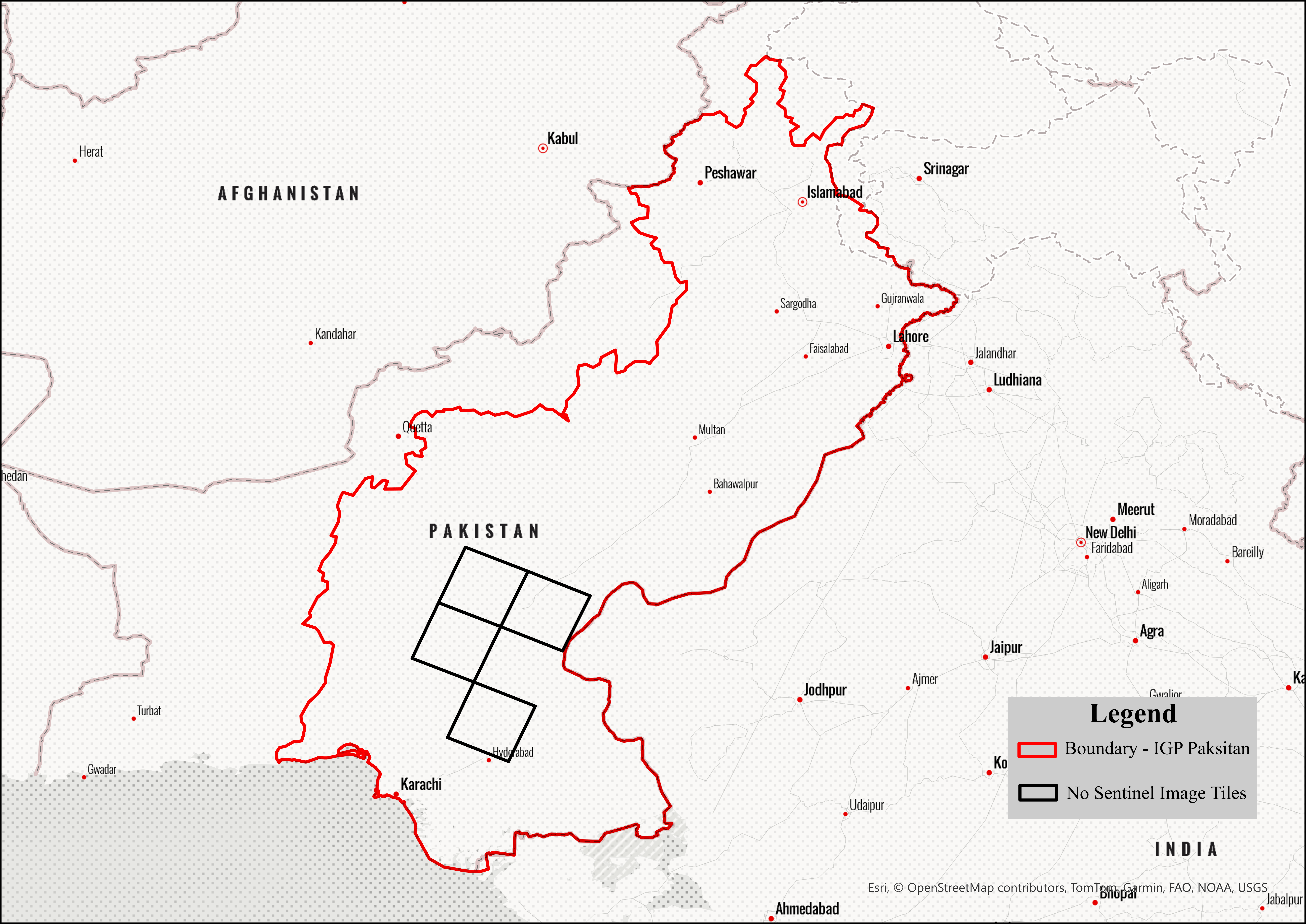}
    \caption{Outlined tiles (black borders) indicate regions where Sentinel-2 imagery was unavailable}
    \label{fig:grids}
\end{figure}